\documentclass[runningheads]{llncs}

\usepackage{graphicx}
\usepackage{amsmath,amssymb,mathtools,bm}
\usepackage{booktabs,multirow,array}
\usepackage[table]{xcolor}
\usepackage{url}
\usepackage[hidelinks]{hyperref}
\usepackage{enumitem}
\usepackage{pifont}
\usepackage{makecell}
\usepackage{caption}
\usepackage{microtype}
\usepackage{wrapfig}
\usepackage{bbding}

\definecolor{bestcell}{gray}{0.90}
\newcommand{\cmark}{\ding{51}}
\newcommand{\xmark}{\ding{55}}
\newcommand{\R}{\mathbb{R}}
\newcommand{\vect}[1]{\boldsymbol{#1}}

\begin{document}

\title{Hull First, Wake Second: Wake-Reliance Suppression for Robust Maritime Vessel Detection}
\titlerunning{Hull First, Wake Second}

\author{
	Yefan Wang\inst{1}\textsuperscript{\ensuremath{\ddagger}}
	\and
	Xingyu Wang\inst{2}\textsuperscript{\ensuremath{\ddagger}}
	\and
	Ruibiao Zhu\inst{3}\Envelope
	\and
	Yusen Wu\inst{4}
}

\authorrunning{Yefan Wang et al.}

\institute{
	University of Shanghai for Science and Technology,
	Shanghai, China\\
	\email{yefanwang88@gmail.com}
	\and
	University of Science and Technology Liaoning,
	Liaoning, China\\
	\email{120243502084@stu.ustl.edu.cn}
	\and
	School of Computing, College of Systems and Society, The Australian National University,
	ACT, Australia\\
	\email{ruibiao.zhu@anu.edu.au}
	\and
	Fujian University of Technology,
	Fujian, China\\
	\email{3231319130@smail.fjut.edu.cn}
}

\maketitle

\begingroup
\renewcommand{\thefootnote}{\ensuremath{\ddagger}}
\footnotetext{Yefan Wang and Xingyu Wang contributed equally to this work.}
\endgroup

\begin{abstract}
	Maritime vessel detectors often face scenes where hulls are small, low-contrast, or blurred, while wakes are longer and easier to detect. This creates a wake-reliance problem: detectors may miss slow or stationary vessels with weak wakes, or produce false positives on wake-like water clutter. We propose HullWake, a hull-first wake-second framework for robust maritime vessel detection. HullWake separates proposal-centered hull evidence from directional wake context, extracts wake cues with bidirectional proposal-anchored corridors, and suppresses wake-dominant predictions through wake response supervision, wake-attenuated consistency, wake-only confidence suppression, and hull--wake decorrelation. We also introduce a wake-oriented evaluation protocol covering weak/no-wake vessels, wake-like hard negatives, worst-group AP, and confidence drop after wake attenuation. Experiments are conducted on Curated-Wake, a wake-oriented maritime dataset of about 10,000 images curated from Ships/Vessels in Aerial Images, the SMD benchmark, and SeaDronesSee, with newly added detection- and segmentation-level wake annotations. Compared with box-only detectors and mask-supervised segmentation baselines, HullWake improves overall AP, weak/no-wake robustness, wake-like false positives, worst-group AP, and confidence stability after wake attenuation.
\end{abstract}

\section{Introduction}

\begin{wrapfigure}{r}{0.55\textwidth}
	\vspace{-16pt}
	\centering
	\includegraphics[width=0.55\textwidth]{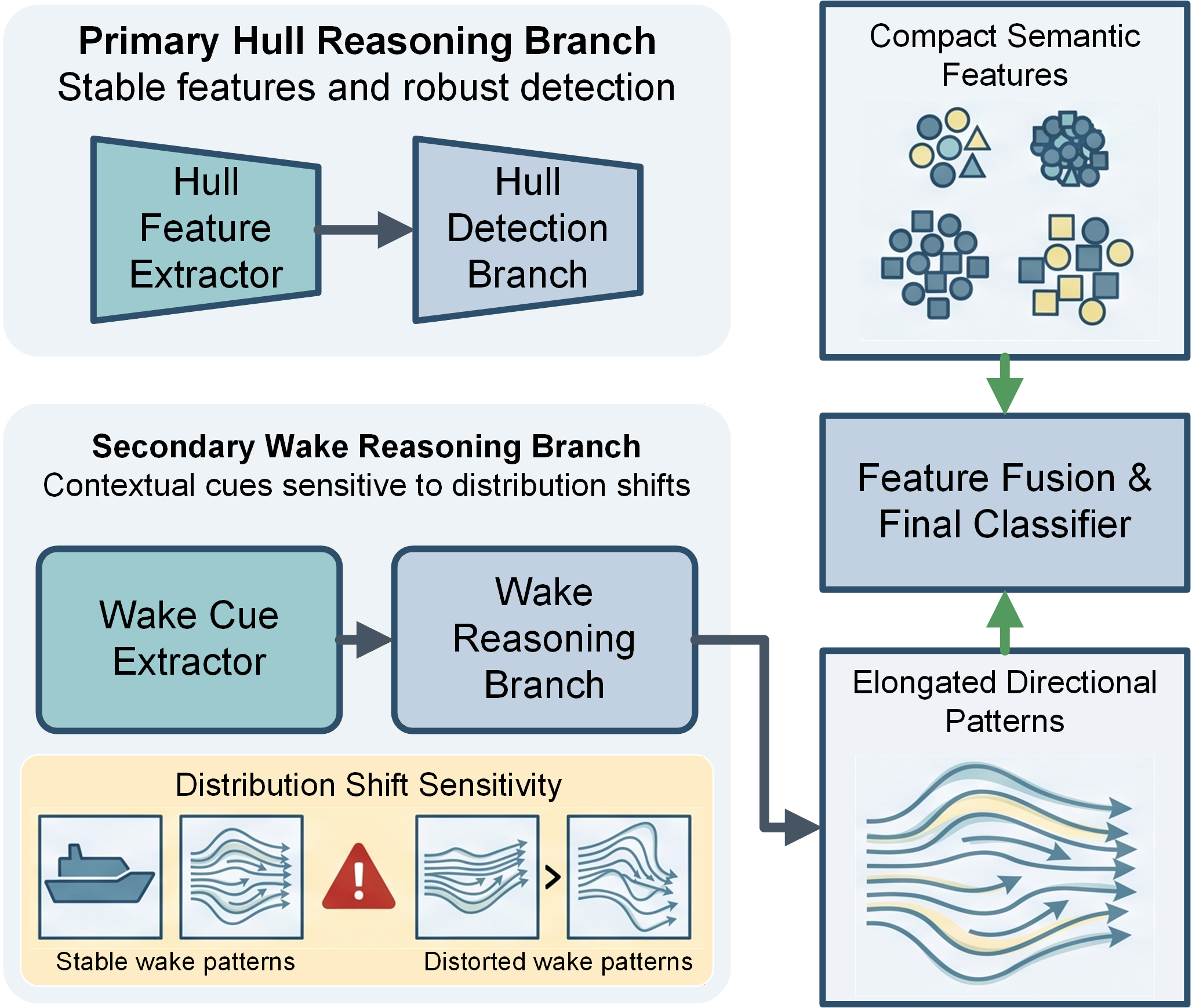}
	\caption{Hull-first and wake-second reasoning.}
	\label{fig:principle}
	\vspace{-16pt}
\end{wrapfigure}

Maritime vessel detection supports coastal surveillance, waterway monitoring, autonomous surface navigation, and remote observation. Unlike generic object detection, water-surface scenes contain unstable context such as waves, reflections, glitter, shoreline clutter, low contrast, scale changes, and motion-induced wakes. Existing maritime datasets and benchmarks have advanced ship and obstacle detection under these conditions~\cite{ships_vessels_aerial_kaggle,9025400,smd_kaggle,7812788,DBLP:conf/wacv/VargaKMZ22}. However, average precision (AP) does not reveal whether a detector verifies the vessel hull or relies on correlated water context. This paper studies \emph{wake reliance}. A moving vessel may leave an elongated wake that is larger and easier to detect than the hull. Wakes are useful because they encode motion and heading, and have long been studied in maritime monitoring and synthetic aperture radar (SAR) imagery~\cite{rs16203775,pichel2004shipwake}. Yet they can also become a shortcut: a detector may treat elongated trailing patterns as evidence of a vessel. This fails for slow, stationary, or weak-wake vessels, and can produce false positives on waves, residual trails, turbulence, reflections, or shoreline traces. Fig.~\ref{fig:principle} summarizes our design premise. Hull evidence is direct and should dominate vessel verification, while wake evidence is contextual and sensitive to speed, sea state, viewpoint, and imaging conditions. We therefore propose HullWake, a hull-first wake-second framework. HullWake extracts proposal-centered hull features for the main detection path, samples directional wake context with bidirectional proposal-anchored corridors, and controls how wake cues enter the final prediction. Wake-dominant decisions are suppressed by wake response supervision, wake-attenuated consistency, wake-only confidence suppression, and hull--wake decorrelation.

We also evaluate the failure mode directly. We build Curated-Wake, a
wake-oriented maritime dataset of about 10,000 images curated from
Ships/Vessels in Aerial Images~\cite{ships_vessels_aerial_kaggle}, the SMD
benchmark~\cite{9025400,smd_kaggle,7812788}, and
SeaDronesSee~\cite{DBLP:conf/wacv/VargaKMZ22}. In addition to vessel boxes, we
add detection-level wake attributes and segmentation-level masks for hull, wake
region, wake-like negative, and water clutter. The evaluation reports standard
AP together with weak/no-wake AP, wake-like false positives, worst-group AP,
and confidence drop after wake attenuation, and compares box-only detectors with
mask-supervised segmentation baselines under the same protocol.

The contributions are: (1) we formulate wake reliance as a measurable shortcut
in maritime vessel detection; (2) we propose a hull-first wake-second detector
with an explicit oriented wake cue extractor; (3) we introduce wake response
supervision and three wake-reliance suppression objectives to keep wake
auxiliary rather than dominant; and (4) we provide a wake-oriented evaluation
protocol beyond overall AP.

\section{Related Work}

\subsection{Maritime Vessel and Water-Surface Detection}
Vision-based maritime perception covers ship detection, obstacle detection, and
water-surface scene understanding. Ships/Vessels in Aerial Images provides a
publicly accessible aerial ship detection dataset with box annotations
~\cite{ships_vessels_aerial_kaggle}. Maritime video surveys and benchmarks
summarize detection and tracking challenges caused by reflections, waves, small
targets, and moving cameras~\cite{9025400,smd_kaggle,7812788}. SeaDronesSee
provides maritime scenes for detecting humans and objects in open water
~\cite{DBLP:conf/wacv/VargaKMZ22}. These datasets mainly report object-level
detection or tracking performance. Our work uses them from a different angle:
whether vessel confidence comes from the hull or from correlated wake context.

\subsection{Ship Wake Modeling and Baselines}
Ship wake is a useful maritime cue because it reflects vessel motion, direction, and sometimes speed. Classical SAR studies use wakes for ship detection and motion analysis~\cite{pichel2004shipwake}. Recent reviews show that wake detection remains relevant in satellite maritime monitoring, especially for small or non-cooperative vessels~\cite{rs16203775}. These works usually treat wake as positive evidence. In this paper, the target is still the vessel hull: wake is modeled explicitly, but its influence is regularized to avoid wake-only decisions.

We use two baseline groups. The first group contains box-level detectors:
Faster R-CNN with FPN~\cite{DBLP:conf/cvpr/LinDGHHB17,DBLP:conf/nips/RenHGS15},
Cascade R-CNN~\cite{DBLP:conf/cvpr/CaiV18},
RetinaNet~\cite{DBLP:conf/iccv/LinGGHD17},
FCOS~\cite{DBLP:conf/iccv/TianSCH19}, YOLO11-m~\cite{yolo11_ultralytics}, and
LSKNet~\cite{10377315}. The second group contains mask-supervised segmentation
models, Mask2Former~\cite{9878483} and PIDNet~\cite{10203574}. These baselines test whether stronger box-level detectors or generic mask supervision can reduce
wake reliance without explicit hull--wake separation.

\subsection{Shortcut Learning and Context Bias}
Context often helps recognition, but it can become a shortcut when it correlates
with labels in training and changes at test time. Shortcut learning has been
studied in deep networks~\cite{DBLP:journals/natmi/GeirhosJMZBBW20}, and
related work on invariant learning and right-for-the-right-reasons training
argues against relying only on the easiest predictive cue
~\cite{DBLP:journals/corr/abs-1907-02893,DBLP:conf/ijcai/RossHD17}. Wake
reliance is a concrete instance of this problem in maritime detection: the cue is
physical, visible, and can be intervened on. This allows us to design a targeted
wake representation, suppression objective, and diagnostic evaluation protocol.

\begin{figure}[t]
	\centering
	\includegraphics[width=\textwidth]{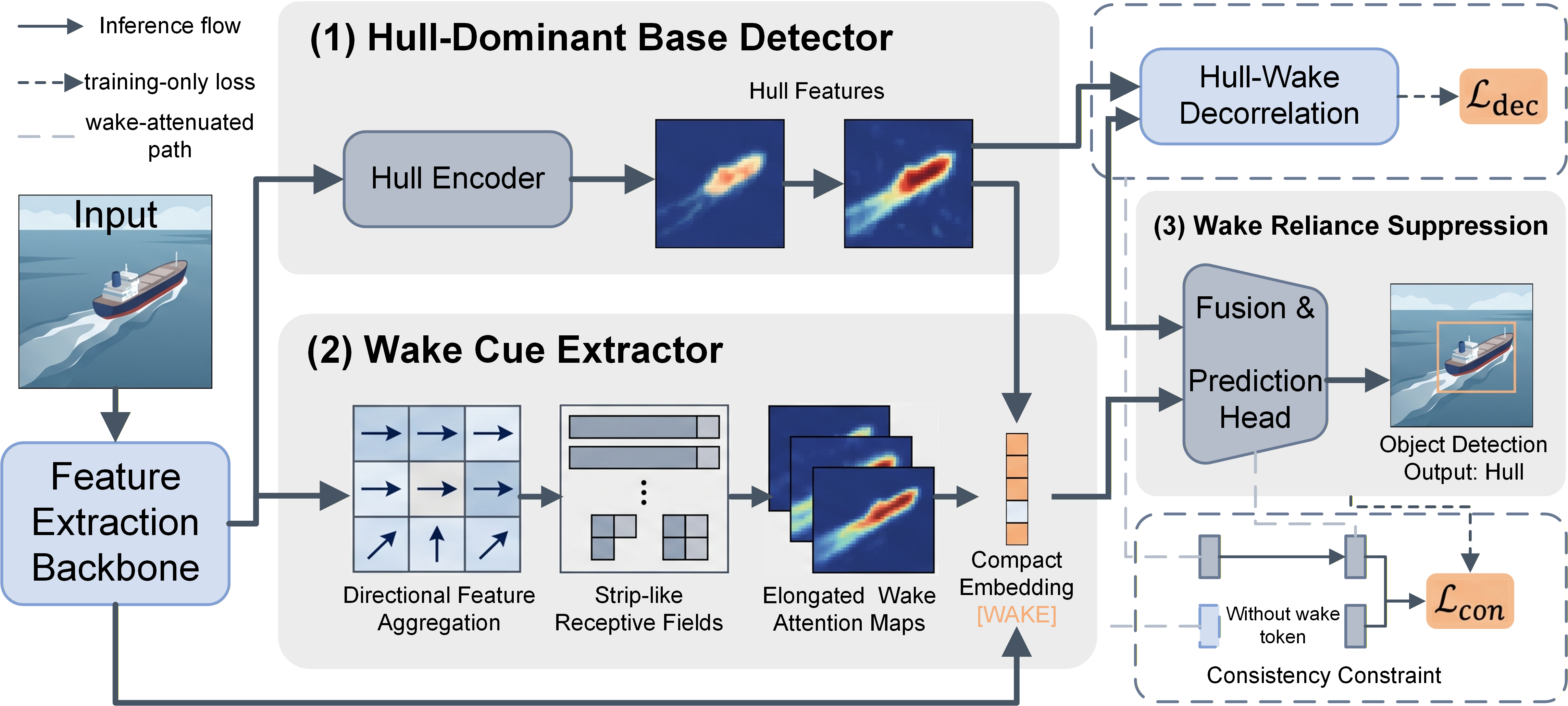}
	\caption{Overview of the proposed HullWake framework. The model follows a hull-first, wake-second design, where wake cues are explicitly extracted and regularized to support rather than dominate vessel detection.}
	\label{fig:framework}
\end{figure}

\section{Method}

\subsection{Problem Formulation and Framework Overview}

Following the hull-first, wake-second principle in Fig.~\ref{fig:principle}, we decompose proposal evidence into hull and wake components. Given a maritime image $x\in\R^{H\times W\times 3}$ and vessel annotations $\mathcal{Y}=\{(b_j^{\mathrm{gt}},y_j)\}_{j=1}^{N}$, where $b_j^{\mathrm{gt}}$ and $y_j$ denote the ground-truth box and class label, a detector predicts candidate boxes $\{b_i\}_{i=1}^{M}$, classification scores $\{s_i\}$, and refined boxes $\{\hat b_i\}$. In our curated wake-oriented dataset, vessel boxes are inherited from the source datasets when available and manually added otherwise. Each vessel instance is further assigned a wake attribute, and segmentation masks are annotated for hull, wake region, wake-like negative, and water clutter regions. For each proposal $b_i$, we use a hull-oriented representation $\vect{z}^{\mathrm{h}}_i$ and a wake-oriented representation $\vect{z}^{\mathrm{w}}_i$. The hull feature provides the main evidence for vessel existence and localization, while the wake feature is used as auxiliary context. We define wake reliance as a conditional shortcut. A detector is wake-reliant if its prediction changes sharply when wake evidence is weakened, or if it assigns high vessel confidence to wake-like water patterns without a visible hull. Ideally, vessel verification should satisfy $p(y_i=1\mid \vect{z}^{\mathrm{h}}_i,\vect{z}^{\mathrm{w}}_i) \approx p(y_i=1\mid \vect{z}^{\mathrm{h}}_i,\operatorname{Att}(\vect{z}^{\mathrm{w}}_i))$, where $\operatorname{Att}(\cdot)$ denotes wake attenuation or removal. The goal is not to remove wake cues, but to prevent them from becoming the main evidence.

Fig.~\ref{fig:framework} shows the architecture of HullWake. Solid arrows denote the inference path, and dashed arrows denote training-only regularization. The framework contains four parts: a hull-dominant detector for proposal-centered evidence, an oriented wake cue extractor for directional trailing context, a controlled fusion head for bounded wake use, and wake-reliance losses for consistency, wake-only suppression, and hull--wake decorrelation.

\subsection{Hull-Dominant Base Detector}

A backbone--neck network extracts a feature pyramid $\mathcal{F}=\{F^{(l)}\}_{l=1}^{L}, F^{(l)}\in\R^{H_l\times W_l\times C_l}$. For each proposal $b_i$, we obtain a proposal-centered feature by RoIAlign~\cite{DBLP:conf/iccv/HeGDG17}: $\vect{r}_i=\operatorname{RoIAlign}(\mathcal{F},b_i) \in\R^{K\times K\times C}$. The hull encoder maps it to $\vect{z}^{\mathrm{h}}_i=\phi_{\mathrm{h}}(\vect{r}_i)\in\R^{d}$. The hull branch predicts $s_i^{\mathrm{h}}=g_{\mathrm{cls}}(\vect{z}^{\mathrm{h}}_i)$ and $\hat b_i=g_{\mathrm{reg}}(\vect{z}^{\mathrm{h}}_i)$, where $g_{\mathrm{cls}}$ and $g_{\mathrm{reg}}$ denote the classification and box-regression heads, respectively. Box regression is kept on the hull feature so that localization is tied to the vessel body rather than trailing water patterns.

\subsection{Oriented Wake Cue Extractor}

For proposal $b_i$, let $\vect{c}_i=(u_i,v_i)$ be its center and $a_i$ its long-side scale. We predict a coarse hull orientation from the hull feature: $\theta_i=g_{\theta}(\vect{z}^{\mathrm{h}}_i), \theta_i\in[-\pi,\pi)$. The parallel and perpendicular directions are $\vect{e}_{\parallel}(\theta_i) = [\cos\theta_i,\sin\theta_i]^{\top}, \vect{e}_{\perp}(\theta_i) = [-\sin\theta_i,\cos\theta_i]^{\top}$. Since monocular hull appearance may have bow--stern ambiguity, we use two candidate trailing corridors:
\begin{equation}
	\begin{split}
		\Omega_i^{+}=
		\{\vect{p}=\vect{c}_i-\alpha\vect{e}_{\parallel}(\theta_i)
		+\beta\vect{e}_{\perp}(\theta_i)
		\mid 0\leq \alpha\leq \ell_i,\ |\beta|\leq w_i/2\},\\
		\Omega_i^{-}=
		\{\vect{p}=\vect{c}_i+\alpha\vect{e}_{\parallel}(\theta_i)
		+\beta\vect{e}_{\perp}(\theta_i)
		\mid 0\leq \alpha\leq \ell_i,\ |\beta|\leq w_i/2\},
	\end{split}
	\label{eq:bidirectional_corridor}
\end{equation}
where $\ell_i=a_i\sigma(g_{\ell}(\vect{z}^{\mathrm{h}}_i))\ell_{\max}, w_i=a_i\sigma(g_{w}(\vect{z}^{\mathrm{h}}_i))w_{\max}$. This avoids using ground-truth heading annotations.

For each corridor $\Omega_i^d$, $d\in\{+,-\}$, feature points are sampled by bilinear interpolation and aggregated by directional attention:
\begin{equation}
	\vect{z}_{i,d}^{\mathrm{w}}
	=
	\sum_{\vect{p}\in\Omega_i^d}
	A_{i,d}(\vect{p})\,\psi(F(\vect{p})).
\end{equation}
The attention weight is
\begin{equation}
	A_{i,d}(\vect{p})
	=
	\frac{
		\exp(q_i^{\top}k(\vect{p})+\rho_d(\vect{p}))
	}{
		\sum_{\vect{p}'\in\Omega_i^d}
		\exp(q_i^{\top}k(\vect{p}')+\rho_d(\vect{p}'))
	},
	\label{eq:directional_attention}
\end{equation}
where $q_i=W_q\vect{z}^{\mathrm{h}}_i$, $k(\vect{p})=W_kF(\vect{p})$, and $\rho_d(\vect{p})$ is a directional position bias:
\begin{equation}
	\rho_d(\vect{p})
	=
	-\eta_{\perp}
	\frac{|\langle \vect{p}-\vect{c}_i,\vect{e}_{\perp}\rangle|}{w_i+\epsilon}
	+
	\eta_{\parallel}
	\frac{\langle \vect{p}-\vect{c}_i,\vect{e}_{d}\rangle}{\ell_i+\epsilon},
	\label{eq:directional_bias}
\end{equation}
where $\vect{e}_{+}=-\vect{e}_{\parallel}$ and $\vect{e}_{-}= \vect{e}_{\parallel}$. The bias favors elongated trailing structures and suppresses off-axis texture.

The two directional descriptors are fused as $\gamma_i = \sigma(g_{\gamma}([\vect{z}_{i,+}^{\mathrm{w}}, \vect{z}_{i,-}^{\mathrm{w}},\vect{z}_i^{\mathrm{h}}])), \vect{z}_i^{\mathrm{w}} = \gamma_i\vect{z}_{i,+}^{\mathrm{w}} + (1-\gamma_i)\vect{z}_{i,-}^{\mathrm{w}}$. The extractor also predicts a soft wake response map $M_i^{\mathrm{w}}(\vect{p}) = \sigma(h_{\mathrm{w}}(F(\vect{p}))), \vect{p}\in\Omega_i^+\cup\Omega_i^- $. In the curated dataset, annotated wake-region masks are used to supervise this response map, while wake-like negative and water-clutter masks are treated as non-wake regions for hard-negative analysis.

\subsection{Controlled Hull--Wake Fusion}

Consistent with Fig.~\ref{fig:principle}, wake is not used as an independent decision source. As shown in Fig.~\ref{fig:framework}, the wake token enters the detector through a bounded fusion head. We use $\alpha_i=\sigma(g_{\alpha}([\vect{z}_i^{\mathrm{h}}, \vect{z}_i^{\mathrm{w}}])), \vect{z}_i^{\mathrm{f}}= \vect{z}_i^{\mathrm{h}} + \alpha_i W_{\mathrm{w}}\vect{z}_i^{\mathrm{w}} $. The final classification score is $s_i=g_{\mathrm{f}}(\vect{z}_i^{\mathrm{f}})$, while box regression remains predicted from $\vect{z}_i^{\mathrm{h}}$. Thus wake can adjust confidence, but not replace hull-based localization.

\subsection{Wake-Attenuated Consistency}

To test whether a prediction depends on wake evidence, we construct a wake-attenuated proposal feature: $\tilde{\vect{r}}_i = \vect{r}_i \odot (1-\lambda_{\mathrm{att}}\uparrow M_i^{\mathrm{w}})$, where $\uparrow$ resizes the wake response to the RoI resolution and $\lambda_{\mathrm{att}}\in[0,1]$ controls attenuation strength. The attenuated hull descriptor is $\tilde{\vect{z}}_i^{\mathrm{h}} = \phi_{\mathrm{h}}(\tilde{\vect{r}}_i)$. The attenuated fused feature is $\tilde{\vect{z}}_i^{\mathrm{f}} = \tilde{\vect{z}}_i^{\mathrm{h}} + \operatorname{sg}(\alpha_i) W_{\mathrm{w}}\operatorname{sg}(\vect{z}_i^{\mathrm{w}})$, where $\operatorname{sg}(\cdot)$ stops gradients. For positive proposals, predictions before and after attenuation should remain close: $\mathcal{L}_{\mathrm{cons}} = \frac{1}{|\mathcal{P}|} \sum_{i\in\mathcal{P}} D_{\mathrm{KL}} \left( p_i\;\middle\|\; \tilde{p}_i \right)$, where $D_{\mathrm{KL}}$ is the Kullback--Leibler divergence, $p_i=\operatorname{softmax}(g_{\mathrm{f}}(\vect{z}_i^{\mathrm{f}}))$, and $\tilde{p}_i=\operatorname{softmax}(g_{\mathrm{f}}(\tilde{\vect{z}}_i^{\mathrm{f}}))$. See lower path in Fig.~\ref{fig:framework}.

\subsection{Wake-Only Confidence Suppression}

We use two auxiliary verifiers to measure hull-only and wake-only confidence: $s_i^{\mathrm{h-only}}=g_{\mathrm{h}}(\vect{z}_i^{\mathrm{h}}), s_i^{\mathrm{w-only}}=g_{\mathrm{w}}(\vect{z}_i^{\mathrm{w}})$. For positive proposals, hull-only confidence should exceed wake-only confidence by margin $m$:
\begin{equation}
	\mathcal{L}_{\mathrm{dom}}^{+}
	=
	\frac{1}{|\mathcal{P}|}
	\sum_{i\in\mathcal{P}}
	\max(0,m+s_i^{\mathrm{w-only}}-s_i^{\mathrm{h-only}}).
	\label{eq:ldompos}
\end{equation}
For negative proposals, especially proposals overlapping wake-like negative or water-clutter regions, wake-only confidence should be low:
\begin{equation}
	\mathcal{L}_{\mathrm{dom}}^{-}
	=
	\frac{1}{|\mathcal{N}|}
	\sum_{i\in\mathcal{N}}
	\operatorname{BCE}(s_i^{\mathrm{w-only}},0),
	\label{eq:ldomneg}
\end{equation}
where $\operatorname{BCE}$ is binary cross-entropy, and $\mathcal{P}$ and $\mathcal{N}$ are positive and negative proposals. The dominance loss is $\mathcal{L}_{\mathrm{dom}} = \mathcal{L}_{\mathrm{dom}}^{+} + \lambda_{\mathrm{neg}}\mathcal{L}_{\mathrm{dom}}^{-}$. This prevents wake-only evidence from becoming sufficient for vessel verification.

\subsection{Hull--Wake Decorrelation}

The dominance loss acts on scores. To separate the feature spaces, we add a normalized hull--wake decorrelation loss. Let $\vect{\mu}_{\mathrm{h}}$ and $\vect{\mu}_{\mathrm{w}}$ be the mini-batch means of hull and wake descriptors. For $N$ proposals in the batch,
\begin{equation}
	\mathcal{L}_{\mathrm{dec}}
	=
	\frac{1}{N}
	\sum_{i=1}^{N}
	\left(
	\frac{
		(\vect{z}_i^{\mathrm{h}}-\vect{\mu}_{\mathrm{h}})^{\top}
		(\vect{z}_i^{\mathrm{w}}-\vect{\mu}_{\mathrm{w}})
	}{
		\|\vect{z}_i^{\mathrm{h}}-\vect{\mu}_{\mathrm{h}}\|_2
		\|\vect{z}_i^{\mathrm{w}}-\vect{\mu}_{\mathrm{w}}\|_2+\epsilon
	}
	\right)^2 .
	\label{eq:decorrelation}
\end{equation}
This term penalizes linear dependence between hull and wake descriptors without forcing wake to be ignored. It corresponds to the upper dashed path in Fig.~\ref{fig:framework}.

\begin{table}[t]
	\caption{Source datasets and added annotations.}
	\label{tab:dataset}
	\centering
	\small
	\resizebox{0.95\linewidth}{!}{
		\begin{tabular}{lcccc}
			\toprule
			Source dataset & Role & Selected images & Original annotation & Added labels \\
			\midrule
			Ships/Vessels in Aerial Images~\cite{ships_vessels_aerial_kaggle}
			& Main source
			& $>$3k
			& Ship boxes
			& Wake attr., masks \\
			SMD benchmark~\cite{9025400,smd_kaggle,7812788}
			& Maritime source
			& $>$3k
			& Object boxes
			& Wake attr., masks \\
			SeaDronesSee~\cite{DBLP:conf/wacv/VargaKMZ22}
			& Cross-view source
			& $>$3k
			& Boxes / tracks
			& Wake attr., masks \\
			\midrule
			Curated dataset
			& Evaluation set
			& $\sim$10k
			& Reused or added hull boxes
			& Detection + segmentation \\
			\bottomrule
		\end{tabular}
	}
\end{table}

\begin{table}[t] \caption{Main results on Curated-Wake.} \label{tab:main} \centering \small \resizebox{0.98\linewidth}{!}{ \begin{tabular}{llccccccc} \toprule Method & Backbone & AP & AP$_{50}$ & AP$_{50:95}$ & AP$_{\mathrm{NoWake}}$ & FP$_{\mathrm{WakeLike}}\downarrow$ & WG-AP & $\Delta_{\mathrm{wake}}\downarrow$ \\ \midrule Faster R-CNN~\cite{DBLP:conf/nips/RenHGS15} & R50-FPN & $54.6{\pm}0.3$ & $79.8{\pm}0.4$ & $51.2{\pm}0.3$ & $43.2{\pm}0.5$ & $118{\pm}4$ & $41.7{\pm}0.4$ & $0.226{\pm}0.011$ \\ Cascade R-CNN~\cite{DBLP:conf/cvpr/CaiV18} & R50-FPN & $56.1{\pm}0.4$ & $81.0{\pm}0.3$ & $52.7{\pm}0.4$ & $44.8{\pm}0.6$ & $112{\pm}5$ & $43.0{\pm}0.5$ & $0.214{\pm}0.010$ \\ RetinaNet~\cite{DBLP:conf/iccv/LinGGHD17} & R50-FPN & $51.9{\pm}0.5$ & $77.2{\pm}0.5$ & $48.5{\pm}0.4$ & $40.6{\pm}0.7$ & $131{\pm}6$ & $39.4{\pm}0.6$ & $0.239{\pm}0.013$ \\ FCOS~\cite{DBLP:conf/iccv/TianSCH19} & R50-FPN & $53.4{\pm}0.4$ & $78.5{\pm}0.4$ & $50.0{\pm}0.4$ & $42.1{\pm}0.6$ & $124{\pm}5$ & $40.8{\pm}0.5$ & $0.231{\pm}0.012$ \\ YOLO11-m~\cite{yolo11_ultralytics} & Default & $57.3{\pm}0.3$ & $82.5{\pm}0.4$ & $53.8{\pm}0.3$ & $45.5{\pm}0.5$ & $109{\pm}4$ & $44.1{\pm}0.4$ & $0.207{\pm}0.010$ \\ LSKNet~\cite{10377315} & LSKNet-S & $58.1{\pm}0.4$ & $83.3{\pm}0.5$ & $54.6{\pm}0.4$ & $46.7{\pm}0.6$ & $102{\pm}5$ & $45.4{\pm}0.5$ & $0.198{\pm}0.010$ \\ Mask2Former~\cite{9878483} & R50 & $58.9{\pm}0.4$ & $84.0{\pm}0.3$ & $55.7{\pm}0.4$ & $48.6{\pm}0.6$ & $90{\pm}5$ & $47.0{\pm}0.5$ & $0.178{\pm}0.009$ \\ PIDNet~\cite{10203574} & PIDNet-M & $58.2{\pm}0.5$ & $83.5{\pm}0.4$ & $55.0{\pm}0.5$ & $47.9{\pm}0.7$ & $94{\pm}5$ & $46.3{\pm}0.6$ & $0.186{\pm}0.011$ \\ \rowcolor{bestcell} HullWake & R50-FPN & $\mathbf{61.8{\pm}0.3}$ & $\mathbf{86.1{\pm}0.3}$ & $\mathbf{58.7{\pm}0.3}$ & $\mathbf{54.6{\pm}0.4}$ & $\mathbf{62{\pm}3}$ & $\mathbf{52.7{\pm}0.4}$ & $\mathbf{0.128{\pm}0.007}$ \\ \bottomrule \end{tabular} } \end{table}

\subsection{Wake Response Supervision and Overall Objective}

Using the annotated segmentation masks in the curated dataset, the soft wake response map is supervised by the wake-region mask:
\begin{equation}
	\mathcal{L}_{\mathrm{wake}}
	=
	\frac{1}{|\Omega|}
	\sum_{\vect{p}\in\Omega}
	\operatorname{BCE}
	(M_i^{\mathrm{w}}(\vect{p}),M_i^{*}(\vect{p})),
	\label{eq:lwake}
\end{equation}
where $\Omega=\Omega_i^+\cup\Omega_i^-$ and $M_i^{*}$ is the annotated wake-region mask restricted to the sampled corridors. Wake-like negative and water-clutter masks are not treated as wake positives; they are used to sample hard-negative regions and to evaluate false wake reliance. This supervision encourages the wake branch to localize actual wake evidence explicitly rather than absorbing unrelated water clutter into the vessel representation. The final training objective is $\mathcal{L} = \mathcal{L}_{\mathrm{det}} + \beta_{\mathrm{wake}}\mathcal{L}_{\mathrm{wake}} + \beta_{\mathrm{cons}}\mathcal{L}_{\mathrm{cons}} + \beta_{\mathrm{dom}}\mathcal{L}_{\mathrm{dom}} + \beta_{\mathrm{dec}}\mathcal{L}_{\mathrm{dec}}$. Here $\mathcal{L}_{\mathrm{det}}$ is the base detector loss. In the main setting, $\beta_{\mathrm{wake}}>0$ because wake-region masks are available in the curated dataset. The remaining terms enforce wake-attenuated consistency, suppress wake-only confidence, and decorrelate hull and wake descriptors.

\section{Experiments}

\subsection{Datasets and Diagnostic Protocol}
We curate a new wake-oriented maritime dataset from three public sources:
Ships/Vessels in Aerial Images~\cite{ships_vessels_aerial_kaggle}, the SMD
benchmark~\cite{9025400,smd_kaggle,7812788}, and
SeaDronesSee~\cite{DBLP:conf/wacv/VargaKMZ22}. From each source, we select more
than 3,000 images, resulting in about 10,000 images in total. Existing vessel
or hull annotations are reused when available, and missing hull annotations are
added manually. On this curated dataset, we add two types of diagnostic labels.

\begin{wrapfigure}{r}{0.5\textwidth}
	\centering
	\includegraphics[width=0.5\textwidth]{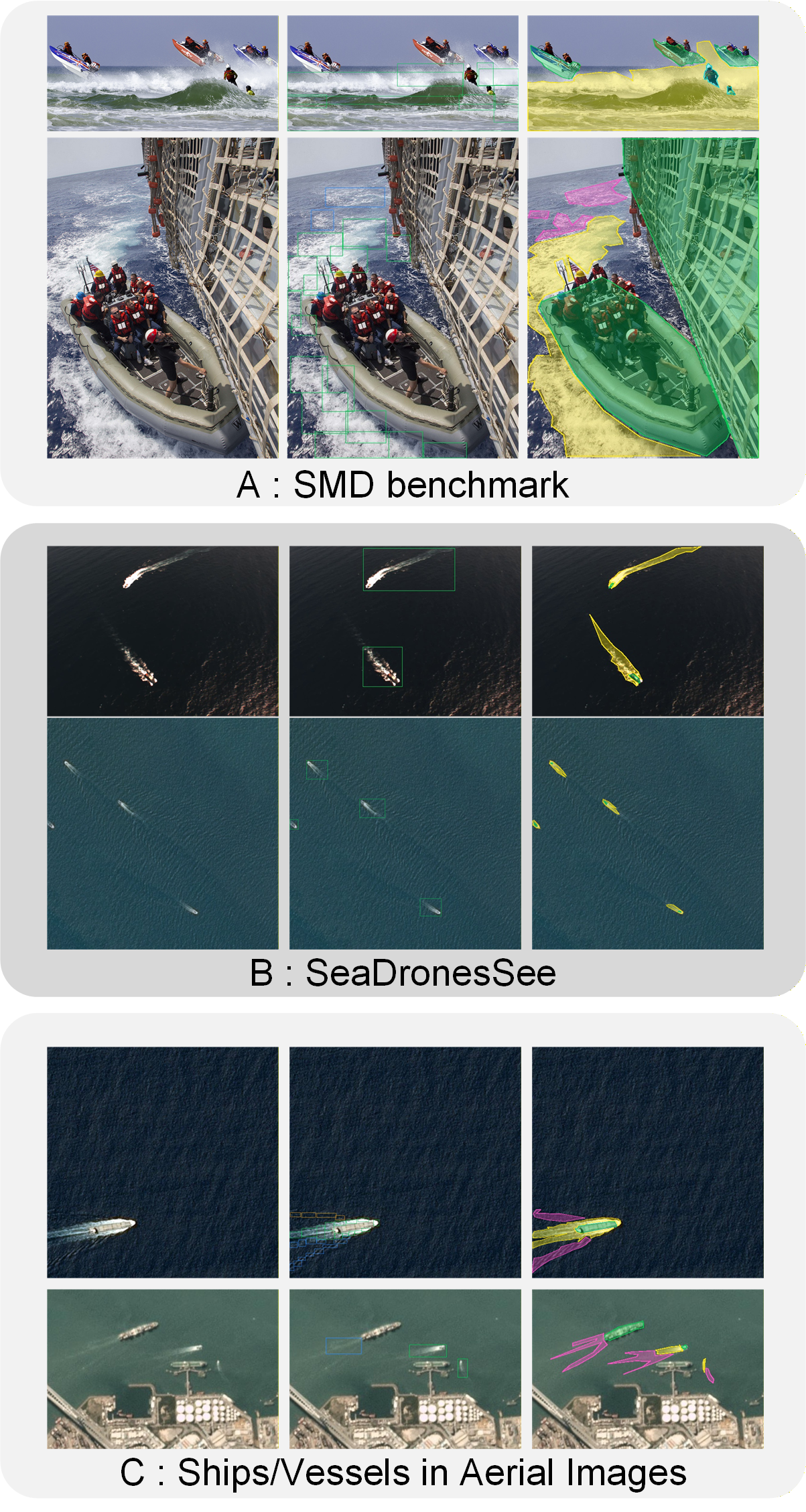}

	\caption{Examples of detection- and segmentation-level wake annotations.}
	\label{fig:diagnostic}
	\vspace{-16pt}
\end{wrapfigure}

For detection, each vessel instance is assigned one wake attribute: clear wake,
weak/no wake, or ambiguous. Ambiguous cases are kept for standard AP but
excluded from group-wise AP. For segmentation, we annotate hull, wake region,
wake-like negative, and water clutter regions.

Fig.~\ref{fig:diagnostic} shows examples of the added labels. Group A is sampled
from the SMD benchmark, group B from SeaDronesSee, and group C from
Ships/Vessels in Aerial Images. In the detection examples, green boxes denote
clear-wake vessels, blue boxes denote weak/no-wake vessels, and yellow boxes
denote ambiguous cases. In the segmentation examples, green masks denote hulls,
yellow masks denote wake regions, magenta masks denote wake-like negatives, and
cyan masks denote water clutter. Water clutter denotes confusing non-wake water
patterns, such as the person mixed with wave clutter in the third image of the
first row in group A. Table~\ref{tab:dataset} summarizes the source datasets and
the added annotations; these labels make it possible to test whether vessel
predictions rely on hull evidence or correlated wake and clutter cues. More importantly, the annotations support condition-wise evaluation across clear-wake, weak/no-wake, and ambiguous cases. 

\subsection{Implementation, Evaluation Metrics, and Results}

\paragraph{Metrics.}

We report AP, AP$_{50}$, and AP$_{50:95}$ following the standard precision--recall definition $\mathrm{AP}=\int_0^1 p(r)\,dr$. To measure wake reliance, we use AP$_{\mathrm{NoWake}}$, FP$_{\mathrm{WakeLike}}$, WG-AP, and $\Delta_{\mathrm{wake}}$, where $\mathrm{WG\text{-}AP}=\min_{g\in\mathcal{G}}\mathrm{AP}_g$ and $\Delta_{\mathrm{wake}}=\frac{1}{|\mathcal{P}|}\sum_{i\in\mathcal{P}} (s_i-\tilde{s}_i)$. AP$_{\mathrm{NoWake}}$ is evaluated on weak/no-wake vessels, FP$_{\mathrm{WakeLike}}$ counts false positives on wake-like water patterns, and ambiguous cases are excluded from group-wise AP.

Unless otherwise stated, HullWake is built on Faster R-CNN with ResNet-50-FPN~\cite{DBLP:conf/cvpr/HeZRS16}. All results are measured on Curated-Wake, whose sources and added annotations are summarized in Table~\ref{tab:dataset}. Box-level detectors, including Faster R-CNN~\cite{DBLP:conf/nips/RenHGS15}, Cascade R-CNN~\cite{DBLP:conf/cvpr/CaiV18}, RetinaNet~\cite{DBLP:conf/iccv/LinGGHD17}, FCOS~\cite{DBLP:conf/iccv/TianSCH19}, YOLO11-m~\cite{yolo11_ultralytics}, and LSKNet~\cite{10377315}, are trained with the reused or newly added vessel/hull boxes. 

\begin{figure}[t]
	\centering
	
	\begin{minipage}[t]{0.5\textwidth}
		\vspace{0pt}
		\centering
		\includegraphics[width=\linewidth]
		{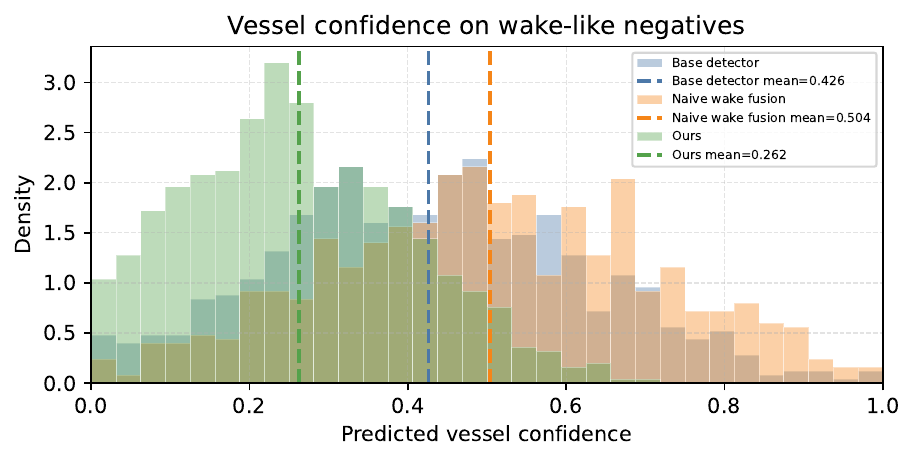}
		
		\vspace{1mm}
		
		\includegraphics[width=\linewidth]
		{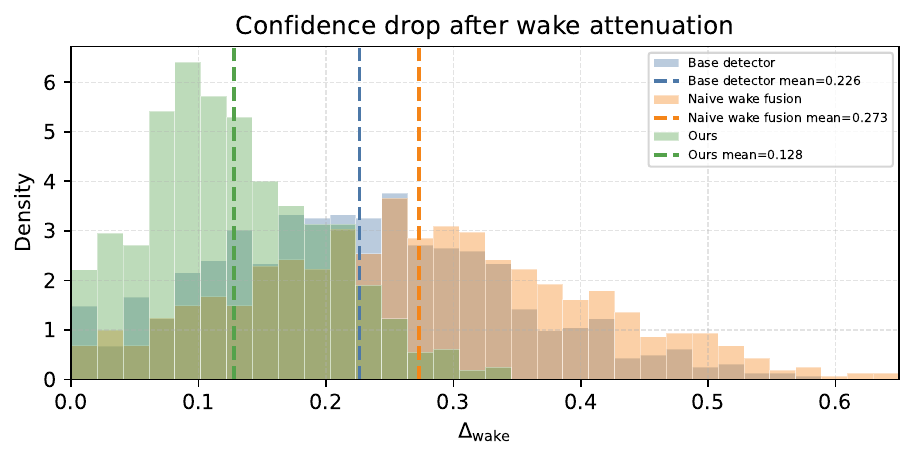}
		\caption{Wake reliance distributions.}
		\label{fig:dist}
	\end{minipage}
	\hfill
	\begin{minipage}[t]{0.46\textwidth}
		\vspace{0pt}
		\centering
		\includegraphics[width=\linewidth]
		{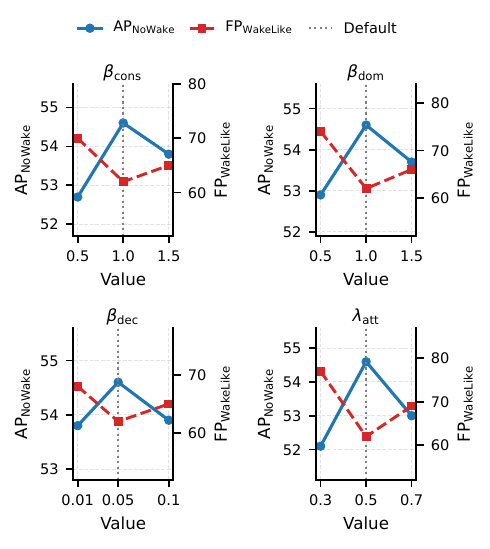}
		\caption{Hyperparameter sensitivity.}
		\label{fig:hparam}
	\end{minipage}
	
\end{figure}

Faster R-CNN, Cascade R-CNN, RetinaNet, and FCOS use images resized with short side 800 and maximum side 1333, and are trained with SGD, momentum 0.9, and weight decay $10^{-4}$. YOLO11-m follows the official Ultralytics setting, and LSKNet follows its remote-sensing detection setting. Mask2Former~\cite{9878483} is trained with the available vessel/hull masks, and its predicted masks are converted to boxes for detection evaluation. PIDNet~\cite{10203574} is a semantic segmentation model; we train it with hull/wake masks, convert connected vessel/hull components to boxes, assign each box the mean foreground probability, and evaluate these boxes under the same detection protocol. Main and ablation results are reported as mean$\pm$std over three random seeds on a single NVIDIA A100 80GB GPU.

For HullWake, the detection loss uses vessel/hull boxes, while the wake branch uses the segmentation labels for wake response supervision and hard-negative analysis. We sample $K=64$ points per direction. The maximum length and width ratios are $\ell_{\max}=3.0$ and $w_{\max}=1.0$, the embedding dimension is $d=256$, and the wake branch uses FPN levels P2--P5. Since wake-region masks are annotated, wake response supervision is enabled in the main setting. We set $\beta_{\mathrm{cons}}=1.0$, $\beta_{\mathrm{dom}}=1.0$, $\beta_{\mathrm{dec}}=0.05$, $\lambda_{\mathrm{neg}}=1.0$, $\lambda_{\mathrm{att}}=0.5$, $m=0.2$, and $\epsilon=0.01$ on the validation split, and vary selected hyperparameters locally around these defaults in Fig.~\ref{fig:hparam} to test stability. Fig.~\ref{fig:hparam} varies these selected hyperparameters around their default values and shows that the robustness trend is stable under moderate changes.

Table~\ref{tab:main} reports the main Curated-Wake results under the same wake-oriented evaluation protocol. Compared with the strongest non-HullWake baseline Mask2Former, HullWake improves AP by 2.9 points and AP$_{\mathrm{NoWake}}$ by 6.0 points, reduces FP$_{\mathrm{WakeLike}}$ from 90 to 62, increases WG-AP by 5.7 points, and lowers $\Delta_{\mathrm{wake}}$ from 0.178 to 0.128. 

\begin{wraptable}{r}{0.5\textwidth}
	\vspace{-18pt}
	\caption{Source-wise robustness gains.}
	\label{tab:source_gain}
	\centering
	\small
	\resizebox{\linewidth}{!}{
		\begin{tabular}{lccc}
			\toprule
			Source & AP$_{\mathrm{NoWake}}$ gain & FP red. & WG-AP gain \\
			\midrule
			SMD & +10.2 & 46.8\% & +9.3 \\
			SeaDronesSee & +8.7 & 43.5\% & +7.8 \\
			Ships/Vessels & +10.5 & 48.1\% & +9.6 \\
			\bottomrule
		\end{tabular}
	}
	\vspace{-18pt}
\end{wraptable}

The box-level and mask-supervised baselines improve ordinary AP over earlier detectors, but they still produce more wake-like false positives and larger confidence drops. Table~\ref{tab:source_gain} further reports gains over Faster R-CNN on the three source-specific subsets, showing that the robustness improvement is consistent across sources rather than dominated by one subset.

\begin{figure}[t]
	\centering
	\includegraphics[width=0.95\linewidth]{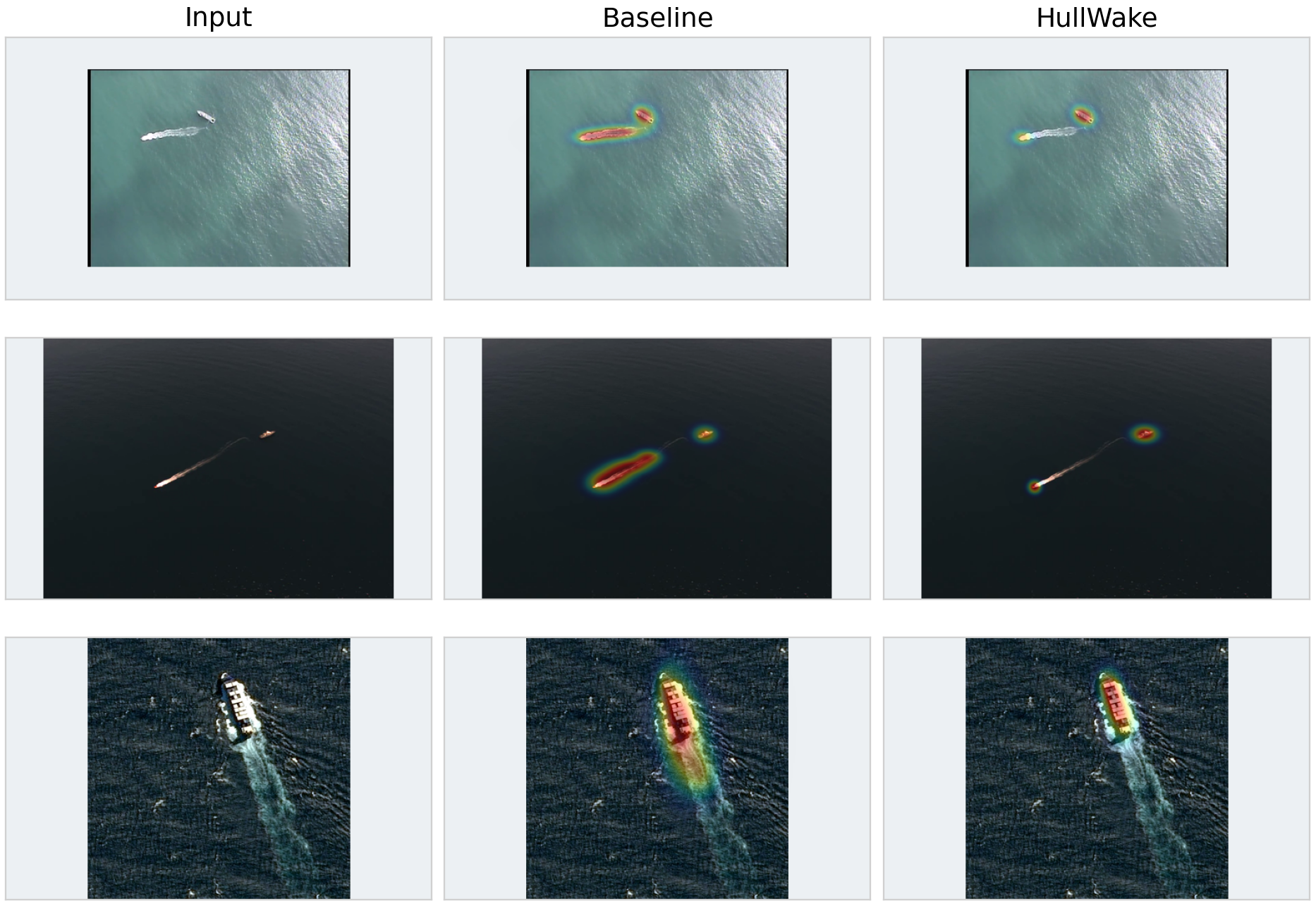}
	\caption{Grad-CAM diagnosis of wake reliance.}
	\label{fig:gradcam}
\end{figure}

Fig.~\ref{fig:dist} provides distribution-level evidence for the wake-reliance metrics. Table~\ref{tab:ablation} uses the same Curated-Wake split and the same Faster R-CNN R50-FPN backbone as the full model. The base detector uses only vessel/hull boxes, while the wake-response rows additionally use the annotated wake-region masks; wake-like negative and water-clutter masks are used for hard-negative analysis and wake-reliance evaluation. Each ``+'' row enables the wake extractor with only the named supervision or regularizer, while Full HullWake enables all components. Naive wake fusion confirms that unconstrained wake context can hurt weak/no-wake robustness and increase wake-like false positives. Wake response supervision gives the strongest single-component gain, wake-only suppression directly reduces FP$_{\mathrm{WakeLike}}$, and the full model gives the best trade-off across AP$_{\mathrm{NoWake}}$, FP$_{\mathrm{WakeLike}}$, WG-AP, and $\Delta_{\mathrm{wake}}$.

\begin{table}[t]
	\caption{Component ablation on Curated-Wake.}
	\label{tab:ablation}
	\centering
	\small
	\resizebox{0.98\linewidth}{!}{
		\begin{tabular}{lccccccccc}
			\toprule
			Variant & Wake Ext. & Wake Sup. & Cons. & Dom. & Dec. & AP$_{\mathrm{NoWake}}$ & FP$_{\mathrm{WakeLike}}\downarrow$ & WG-AP & $\Delta_{\mathrm{wake}}\downarrow$ \\
			\midrule
			Base detector & \xmark & \xmark & \xmark & \xmark & \xmark & $43.2{\pm}0.5$ & $118{\pm}4$ & $41.7{\pm}0.4$ & $0.226{\pm}0.011$ \\
			Naive wake fusion & \cmark & \xmark & \xmark & \xmark & \xmark & $41.9{\pm}0.6$ & $137{\pm}6$ & $40.5{\pm}0.5$ & $0.273{\pm}0.014$ \\
			+ wake response supervision & \cmark & \cmark & \xmark & \xmark & \xmark & $50.6{\pm}0.5$ & $76{\pm}4$ & $48.9{\pm}0.4$ & $0.152{\pm}0.009$ \\
			+ wake-attenuated consistency & \cmark & \xmark & \cmark & \xmark & \xmark & $48.4{\pm}0.6$ & $98{\pm}5$ & $46.7{\pm}0.5$ & $0.181{\pm}0.010$ \\
			+ wake-only suppression & \cmark & \xmark & \xmark & \cmark & \xmark & $49.7{\pm}0.5$ & $82{\pm}4$ & $47.9{\pm}0.4$ & $0.166{\pm}0.009$ \\
			+ hull--wake decorrelation & \cmark & \xmark & \xmark & \xmark & \cmark & $49.1{\pm}0.5$ & $91{\pm}5$ & $47.2{\pm}0.5$ & $0.158{\pm}0.008$ \\
			\rowcolor{bestcell}
			Full HullWake & \cmark & \cmark & \cmark & \cmark & \cmark & $\mathbf{54.6{\pm}0.4}$ & $\mathbf{62{\pm}3}$ & $\mathbf{52.7{\pm}0.4}$ & $\mathbf{0.128{\pm}0.007}$ \\
			\bottomrule
		\end{tabular}
	}
\end{table}

HullWake assigns lower vessel confidence to wake-like negatives and shows a smaller confidence drop after wake attenuation, indicating fewer clutter-induced false positives and weaker dependence on wake evidence. Together with Fig.~\ref{fig:hparam}, the results suggest that the gains come from suppressing wake-dominant evidence rather than from a fragile parameter choice. We further visualize Grad-CAM~\cite{DBLP:conf/iccv/SelvarajuCDVPB17} in Fig.~\ref{fig:gradcam}. For a controlled comparison, Grad-CAM is computed on the same Curated-Wake images for the Faster R-CNN R50-FPN baseline and HullWake. The baseline responses tend to extend to wake or wake-like water patterns, while HullWake produces more hull-centered activations. This qualitative diagnosis is consistent with the lower FP$_{\mathrm{WakeLike}}$ and smaller $\Delta_{\mathrm{wake}}$ in Table~\ref{tab:main}.

\section{Conclusion}
This work contributes to safer maritime perception for coastal surveillance, waterway monitoring, and autonomous surface navigation by reducing wake-driven detection failures. HullWake extracts directional wake context through proposal-anchored corridors, but regularizes the detector so that wake supports rather than dominates vessel verification. The method combines wake response supervision, wake-attenuated consistency, wake-only confidence suppression, and hull--wake decorrelation. We also introduced Curated-Wake, a wake-oriented maritime dataset of about 10,000 images curated from three public sources with added detection- and segmentation-level wake annotations. Experiments against box-level detectors and mask-supervised segmentation baselines show that HullWake improves overall AP, strengthens weak/no-wake robustness, reduces wake-like false positives, increases worst-group AP, and lowers confidence drop after wake attenuation. These results indicate that generic detection or mask supervision alone is not sufficient to remove wake reliance; wake evidence is useful only when its shortcut effect is constrained by hull-centered verification. Future work will extend the wake-reliance analysis to broader maritime conditions.

\subsubsection*{Competing Interests}
The authors declare no competing interests relevant to this work.
	
\bibliographystyle{splncs04}
\bibliography{bibliography}
	
\end{document}